\documentclass{ecai} 

\usepackage{latexsym}
\usepackage{amsthm}
\usepackage{enumitem}

\usepackage{amsmath}
\usepackage{amssymb}
\usepackage{subcaption}
\usepackage{mathptmx}
\usepackage{booktabs}
\usepackage{multirow}
\usepackage{array}
\usepackage{graphicx}
\usepackage{latexsym}
\usepackage{tikz}
\usepackage{xspace}
\usepackage{pgfplots}
\usetikzlibrary{decorations.pathreplacing}
\usepackage{makecell}

\usepackage{comment}
\excludecomment{excludesection}

\usepackage{ifthen}
\newboolean{isold}
\setboolean{isold}{false}

\usepackage{color}
\definecolor{rblue}{RGB}{65,105,225}
\definecolor{rgreen}{RGB}{46,139,87}
\definecolor{rred}{RGB}{220,20,60}
\definecolor{ryellow}{RGB}{218,165,32}
\definecolor{sgray2}{RGB}{81,91,101}
\definecolor{sgray3}{RGB}{104,115,125}
\definecolor{sgray4}{RGB}{168,173,185}

\newif\iflongversion

\longversiontrue

\newcommand{\cX}{\mathcal{X}}
\newcommand{\cC}{\mathcal{C}}
\newcommand{\cS}{\mathcal{S}}
\newcommand{\R}{\mathbb{R}}

\newcommand{\op}{\textsc{op}\xspace}
\newcommand{\fop}{\textsc{fop}\xspace}
\newcommand{\hfop}{\textsc{hfop}\xspace}
\newcommand{\dhfop}{\textsc{dhfop}\xspace}
\newcommand{\msdhfop}{\textsc{msdhfop}\xspace}

\newtheorem{definition}{Definition}

\newcommand{\BibTeX}{B\kern-.05em{\sc i\kern-.025em b}\kern-.08em\TeX}

\begin{document}


\begin{frontmatter}


\paperid{2648} 


\title{Temporal Fairness in Decision Making Problems}


\author[A]{\fnms{Manuel R.}~\snm{Torres}\thanks{Corresponding Author. Email: manuel.torres@jpmchase.com.}\footnote{Equal contribution.}}
\author[A]{\fnms{Parisa}~\snm{Zehtabi}\footnotemark}
\author[A]{\fnms{Michael}~\snm{Cashmore}} 
\author[A]{\fnms{Daniele}~\snm{Magazzeni}} 
\author[A]{\fnms{Manuela}~\snm{Veloso}} 

\address[A]{J.P. Morgan AI Research}


\begin{abstract}
In this work we consider a new interpretation of fairness in decision making problems.
Building upon existing fairness formulations, we focus on how to reason over fairness from a temporal perspective, taking into account the fairness of a history of past decisions.
After introducing the concept of temporal fairness, we propose three approaches that incorporate temporal fairness in decision making problems formulated as optimization problems.
We present a qualitative evaluation of our approach in four different domains and compare the solutions against a baseline approach that does not consider the temporal aspect of fairness.
\end{abstract}

\end{frontmatter}


\section{Introduction}

Automated decision making is an important part of artificial intelligence with a variety of application areas, from scheduling and resource allocation, to robotics and autonomous vehicles.
Decision making processes typically aim to optimize an overall benefit or cost. 
However, as we strive to make our algorithms and agents more intelligent, it is important to ensure that they also account for ethical considerations such as fairness.
The need for fair algorithms and agents has been widely studied across different areas, such as robotics~\cite{brandao2020fair}, healthcare~\cite{chen2021ethical}, telecommunications~\cite{huaizhou2013fairness}, and resource allocation~\cite{kumar2000fairness}, among others.

Formulating fairness concerns in different domains can be challenging, and it has been the subject of many studies~\cite{xinying2023guide}.
In this paper, we take a new angle to considering fairness in decision making processes.
We build upon previous fairness formulations, and focus on how to reason about fairness from a temporal perspective, accounting for the fairness of a history of past decisions.
We aim to introduce the concept of \textit{``temporal fairness''} into the decision making process, which measures the fairness of solutions throughout time.

\begin{figure}[t]
\centering
\includegraphics[width=.7\columnwidth]{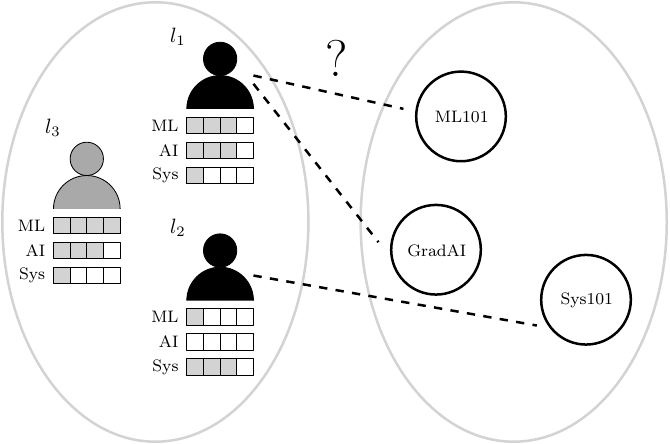}
\caption{Depicts a \textit{university course assignment}, in which courses are to be assigned to lecturers with different levels of expertise in different areas. The lecturer in gray is on sabbatical leave. }
\label{fig:courses domain}
\vspace{6mm}
\end{figure}

\begin{figure*}[t!]
    \begin{subfigure}[c]{0.3\textwidth}
        \centering
        \begin{tabular}{@{}ccccc@{}}
            \multicolumn{5}{c}{Courses lectured in past 4 semesters} \\ \toprule
                 & $t-4$ & $t-3$ & $t-2$ & $t-1$ \\ \midrule
            $l_1$ & 2  & 1.5 & 3 & 2   \\
            $l_2$ & 1  & 1.5 & 0 & 1   \\
            \bottomrule
        \end{tabular}        
        \caption{}
        \label{subfig:unfair previous plans}
        \vspace{3mm}
    \end{subfigure}
    \begin{subfigure}[c]{0.35\textwidth}
    \centering
        
    \begin{tikzpicture}[scale=0.55]
    \begin{axis}[
        xmin=-4, xmax=0,
        ymin=0, ymax=10.5,
        xtick={-4,-3,-2,-1,0},
        xticklabels={$t-4$,$t-3$,$t-2$,$t-1$,$t$},
        ytick={0,1,2,3,4,5,6,7,8,9,10},
        yticklabels={,1,2,3,4,5,6,7,8,9,10},
        grid=both,
        xlabel={Semester},
        ylabel={Cumulative Courses Lectured},
        ylabel style={yshift=-15pt},
        legend style={at={(.98, 0.2)},
                      anchor=north east},
        ]
    
        \addplot[forget plot, darkgray,thick,dashed,mark=*] coordinates {
            (-1,8.5) (0,10)
        };
        \addplot[forget plot, darkgray,thick,dashed,mark=*] coordinates {
            (-1,3.5) (0,5.0)
        };
    
        \addplot[rblue,thick,mark=*] coordinates {
            (-4,2) (-3,3.5) (-2,6.5) (-1,8.5)
        };
    
        \addplot[rred,thick,mark=*] coordinates {
            (-4,1) (-3,2.5) (-2,2.5) (-1,3.5)
        };

    \draw[<->, thick] (300,80) -- (300,40) node[midway, fill=white,inner sep=4pt] {5};

        \legend{$l_1$, $l_2$}
    \end{axis}
    \end{tikzpicture}

    \caption{}
    \label{subfig:fairness debt fair plan}
    \vspace{3mm}
    \end{subfigure}
    \begin{subfigure}[c]{0.35\textwidth}
    \centering
        
    \begin{tikzpicture}[scale=0.55]
    \begin{axis}[
        xmin=-4, xmax=0,
        ymin=0, ymax=10.5,
        xtick={-4,-3,-2,-1,0},
        xticklabels={$t-4$,$t-3$,$t-2$,$t-1$,$t$},
        ytick={0,1,2,3,4,5,6,7,8,9,10},
        yticklabels={,1,2,3,4,5,6,7,8,9,10},
        grid=both,
        xlabel={Semester},
        ylabel={Cumulative Courses Lectured},
        ylabel style={yshift=-15pt},
        legend style={at={(.98, 0.2)},
                      anchor=north east},
        ]
    
        \addplot[forget plot, darkgray,thick,dashed,mark=*] coordinates {
            (-1,8.5) (0,8.5)
        };
        \addplot[forget plot,darkgray,thick,dashed,mark=*] coordinates {
            (-1,3.5) (0,6.5)
        };
    
        \addplot[rblue,thick,mark=*] coordinates {
            (-4,2) (-3,3.5) (-2,6.5) (-1,8.5)
        };
    
        \addplot[rred,thick,mark=*] coordinates {
            (-4,1) (-3,2.5) (-2,2.5) (-1,3.5)
        };

        \legend{$l_1$, $l_2$}
    \end{axis}
    \end{tikzpicture}

        \caption{}
        \label{subfig:fairness debt unfair plan}
    \vspace{3mm}
    \end{subfigure}
    \caption{Depicts a concrete scenario of a university course assignment. Figure~\ref{subfig:unfair previous plans} depicts the scenario in which $l_1$ was assigned a higher lecturing load over the past 4 semesters. Figure~\ref{subfig:fairness debt fair plan} and~\ref{subfig:fairness debt unfair plan} depict the cumulative lecturing load of $l_1$ (blue) and $l_2$ (red) with possible solutions to the current allocation problem (dashed gray). \ref{subfig:fairness debt fair plan} depicts the case where at timestep $t$ a fair allocation is made with both lecturers given the same load ($1.5$ courses each). \ref{subfig:fairness debt unfair plan} depicts the case where an unfair allocation is made and $l_2$ is assigned all the lecturing load (3 courses).}
    \label{fig:lecture domain}
    \vspace{5mm}
\end{figure*}
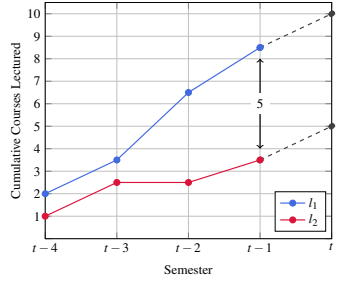
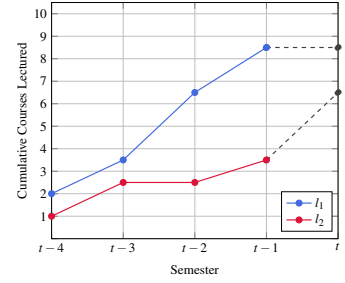

As a motivating example consider the scenario depicted in Figure~\ref{fig:courses domain} where courses must be assigned to a pool of lecturers ($l_1, l_2$ and $l_3$) in semester $t$. Each lecturer is specialized in different areas and the teaching quality of a course is proportional to the expertise of its lecturer (the gray bars below the lecturers depict their expertise on different topics).
Figure~\ref{subfig:unfair previous plans} depicts the number of courses assigned to each lecturer in the past four semesters. Lecturer $l_1$ has received a higher teaching load than $l_2$ over the past four semesters. Regardless of the reasons that have led to the scenario in Figure~\ref{subfig:unfair previous plans}, the reality is that there has been some \textit{``historical unfairness''}. Figure~\ref{subfig:fairness debt fair plan} depicts the cumulative teaching load over time for each of the two available lecturers $l_1$ and $l_2$.

A new course allocation must be made for semester $t$. If an allocation is made that is presently fair, in which both lecturers teach the same number of courses (dashed gray scenario in Figure~\ref{subfig:fairness debt fair plan}), an overall \textit{``temporal unfairness''} remains, with the gap in cumulative lecturing load not reducing. In fact, as depicted in Figure~\ref{subfig:fairness debt unfair plan}, even in the case where an unfair allocation is made for semester $t$ and $l_2$ lectures all courses, there would still exist a gap in the lecturing load.

In this work, we focus on the problem of decision making while accounting for a historical fairness and considering the impact of future decisions in overall temporal fairness.

We first introduce the definition of an optimization problem that reasons over the trade-off between quality and fairness (Section~\ref{sec:fop}). Then we introduce the concept of temporal fairness by including historical fairness into our formulation (Sections~\ref{sec:hfop} and \ref{sec:dhfop}). We then extend this optimization problem to reason over this trade-off while accounting for future predictions and forecasts (Section~\ref{sec:mshfop}). This allows the generation of solutions that may look unfair in the short term, but fairer when analyzed over a longer period of time into the future. We incorporate the notion of temporal fairness via the introduction of a framework for fairness metrics that considers historical solutions.

The main contributions of this paper are: 
(i) introducing the concept of \textit{temporal fairness} in decision making problems, 
(ii) a formulation for addressing historical unfairness from past solutions,
(iii) a formulation for both addressing historical unfairness and considering future historical fairness, and 
(iv) a qualitative evaluation on different domains that examines the differences between solutions generated with and without considering the temporal aspect of fairness.

The remainder of the paper is structured as follows. 
Section~\ref{sec:problem formulation} introduces the formulations of optimization problems that consider fairness, as described above. We then present the qualitative evaluation of our framework in Section~\ref{sec:experimental evaluation}.
Section~\ref{sec:related work} discusses relevant related work, and the paper concludes in Section~\ref{sec:conclusion} with final remarks and a discussion on avenues for future work.

\section{Problem Formulation}
\label{sec:problem formulation}

We consider decision making problems solved by finding a solution that maximizes an objective function while satisfying a set of constraints. First, we consider a decision making problem where no fairness metric is considered. Such problems can be formulated as an Optimization Problem (\op).
\begin{definition}\label{def_OP}
    An \textbf{Optimization Problem ($\mbox{OP}$)} is a tuple $\langle Q, \mathcal{X}, \mathcal{C}\rangle$ where $Q$ is a quality metric, $\mathcal{X}$ is the domain
    for optimization, and $\mathcal{C}$ is the set of constraints. 
\end{definition}
Formally, we define an \op as:
\begin{equation}
    \label{eq:setting 1}
    \begin{array}{ll@{}ll}
    \underset{x\in\mathcal{X}}{\max} & Q(x) & \\
    \ \ \text{s.t.} & \mathcal{C}(x)
\end{array}
\end{equation}
In this setting, the goal is to find a solution $x^*$ that maximizes a given \emph{quality metric} $Q$, while being subject to a set of constraints $\mathcal{C}(x)$.
We let variables $x$ denote the optimization variables of the problem.
Since this formulation only reasons over the quality metric, it is possible that the optimal solutions may be deemed unfair according to some fairness metric.
Moreover, as depicted in Figure~\ref{subfig:unfair previous plans}, this formulation may lead to a fast accumulation of unfair solutions.

\subsection{FOP: Incorporating Fairness}\label{sec:fop}

We now incorporate a fairness metric $F$ into the formulation of the optimization problem. A Fair Optimization Problem (\fop) can be defined as:
\begin{definition}\label{def:fop}
    A \textbf{Fair Optimization Problem ($\mbox{FOP}$)} is a tuple $\langle Q, F, \mathcal{X}, \mathcal{C}, \beta\rangle$ where $F$ is the fairness metric and $\beta\in\R$ is a parameter than controls the trade-off between quality and fairness. The remaining elements follow the original OP.
\end{definition}
Formally, an \fop can be modelled as:
\begin{equation}
    \label{eq:historyless fairness}
    \begin{array}{ll@{}ll}
    \underset{x\in\mathcal{X}}{\max} & Q(x) + \beta F(x) & \\
    \ \ \text{s.t.} & \mathcal{C}(x)
\end{array}
\end{equation}
In general, we will assume that $F$ returns higher values for fair solutions and lower values for unfair solutions.
In practice, it may be convenient for both $Q$ and $F$ to have well-specified ranges, rendering it easier to understand the impact of the parameter $\beta$.
However, the formulation is general and supports arbitrary quality and fairness metrics. Finally, we note that the specification of the fairness metric $F$ may potentially require the introduction/modification of constraints.
In order to keep notation simple, we will continue denoting the set of constraints as before, $\mathcal{C}(x)$.

As an example building upon our previous scenario of the course assignment domain, let us consider a \emph{relative max-min} fairness metric $F^{\text{rmm}}$, which compares the maximum and minimum number of courses lectured by all lecturers, versus the total number of courses lectured during that time.
Formally,
\begin{equation*}
    F^\text{rmm}(x) = 1 - \frac{\max_{i} S_i(x) - \min_j S_j(x)}{S(x)},
\end{equation*}
where $S_i(x)$ is the number of courses lectured by $l_i$ in solution $x$, and $S(x)$ is the total number of courses lectured.
The range of $F^\text{rmm}$ is $[0, 1]$. It is maximized when lecturers get an equal lecturing load, and minimized when one of the lecturers takes the entire load.
While incorporating the new fairness metric in the \fop leads to solutions that are fair according to $F$ (or at least fairer, depending on $\beta$), there may still exist some historical unfairness that remains from previous allocations. 
Figures~\ref{subfig:unfair previous plans} and~\ref{subfig:fairness debt fair plan}  hinted at this, depicting a scenario where scheduling a fair plan at time step $t$ would have maintained the gap of cumulative courses lectured.
\ifthenelse{\boolean{isold}}{
To tackle this issue, we propose to reason over fairness from a temporal perspective.
\begin{definition}\label{TF}
    \textbf{Temporal Fairness} is a metric that considers the fairness of a solution in the context of both historical solutions as well as envisioned future problems.
\end{definition}

We now extend {\fop}s to reason over temporal fairness.
}{
}

\subsection{HFOP: Incorporating Historical Fairness}\label{sec:hfop}

\textsc{fop}s assume a fairness metric $F$ that only reasons over the fairness of a solution $x$.
In order to account for existing historical unfairness, it is thus important to reason over the fairness of a solution $x$ in the context of the history of past solutions $H = (x_{t-T}, \dots, x_{t-1})$, where $x_{t-\Delta}$ is a previous solution from time step $t-\Delta$.

We formalize the notion of such fairness metrics in the following definition.
\begin{definition}
  A \emph{historical fairness} metric $F_H : \cX \to \R$ is a fairness metric for an FOP 
  $\langle Q, F_H, \cX, \cC, \beta\rangle$ where $H \subseteq \cX$ contains solutions satisfying $\cC$.
  We assume $F_\emptyset$ is a fairness metric and write it as $F$.
\end{definition}
The historical fairness metric $F_H$ can be used to control how fast or slow historical unfairness is compensated. Also, as it is not a limitation for the real-world scenarios
we consider in this paper, we assume the time between the historical solutions in $H$
is uniform.

\begin{definition}
  A \textbf{Historical Fair Optimization Problem ($\mbox{HFOP}$)} is an \fop tuple $\langle Q, F_H, H, \mathcal{X}, \mathcal{C},\beta\rangle$ where $F_H$ is a historical fairness metric.
\end{definition}

Formally, we can formulate an \hfop as:
\begin{equation}
    \label{eq:historical fairness}
    \begin{array}{ll@{}ll}
    \underset{x\in\mathcal{X}}{\max} & Q(x) + \beta F_{H}(x) & \\
    \ \ \text{s.t.} & \mathcal{C}(x)
\end{array}
\end{equation}
As before, parameter $\beta$ provides control over the quality/fairness trade-off, with higher values of $\beta$ leading to a faster compensation of historical unfairness.
It is worth highlighting that the optimal solution to an \hfop may actually be an unfair solution from the perspective of a fairness metric $F$. To see this, let us consider an example.

Building upon the \emph{relative max-min} fairness metric previously discussed, we can now consider its historical variant $F_H^{\text{rmm}}$, where we reason instead over the courses lectured across $(H, x)$---the concatenation of historical solutions in $H$ with the new solution $x$.
\begin{equation*}
    F_H^\text{rmm}(x) = 1 - \frac{\max_{i} S_i((H, x)) - \min_j S_j((H, x))}{S((H, x))},
\end{equation*}
where $S_i((H, x))$ is the number of courses lectured by $l_i$ over all solutions in $(H, x)$ and $S((H, x))$ is the total number of courses lectured.

Let's now see how $F^\text{rmm}$ and $F_H^\text{rmm}$ would differ in a concrete scenario, depicted in Table~\ref{tab:compare F and F_H}.
Assume that at time step $t$ we take a solution $x_\text{(1.5, 1.5)}$ assigning an equal load of 1.5 courses to each lecturer.
While $F^\text{rmm}(x_\text{(1.5,1.5)}) = 1$, we have that $F_H^\text{rmm}(x_\text{(1.5,1.5)}) = 1 - \frac{5}{15} \sim 0.67$. Since the allocation given by 
$x_\text{(1.5,1.5)}$ is balanced, the historical max-min gap remains $5$ while the total number of courses becomes 15.
On the other hand, if we take the solution $x_\text{(0,3)}$ assigning all 3 courses to lecturer $l_2$, we would have $F^\text{rmm}(x_\text{(0,3)}) = 0$ and $F_H^\text{rmm}(x_\text{(0,3)}) = 1 - \frac{2}{15} \sim 0.87$ -- since all the lecturing load was assigned to $l_2$, the max-min gap decreases to 2.

\begin{table}
  \vspace{3mm}
    \scriptsize
    \centering
        \caption{Compares $F$ and $F_H$ for different solutions, assuming a history $H$ (repeated from Figure~\ref{subfig:unfair previous plans}). A solution $x_{(m,n)}$ refers to the case where lecturers $l_1$ and $l_2$ are assigned $m$ and $n$ courses, respectively. We observe that, under history $H$, a perfectly fair solution according to $F$ may not be fair according to $F_H$. The solutions that would be picked under $F$ and $F_H$ are in bold.}
    \label{tab:compare F and F_H}
    \begin{tabular}{@{}ccccc@{}}
        \multicolumn{5}{c}{$H$} \\ \midrule
             & $t\!-\!4$ & $t\!-\!3$ & $t\!-\!2$ & $t\!-\!1$ \\ \midrule
        $l_1$ & 2  & 1.5 & 3 & 2   \\
        $l_2$ & 1  & 1.5 & 0 & 1   \\
        \midrule
    \end{tabular}  
    \quad
    \begin{tabular}{@{}cccc@{}}
        $x_t$           & $F(x_t)$    & $F_H(x_t)$ \\ \midrule
        $x_{(1.5,1.5)}$ &  \textbf{1.00}       & 0.67       \\
        $x_{(1,2)}$     &  0.67       & 0.73       \\
        $x_{(0,3)}$     &  0.00        & \textbf{0.87 }      \\ \midrule
    \end{tabular}
\end{table}

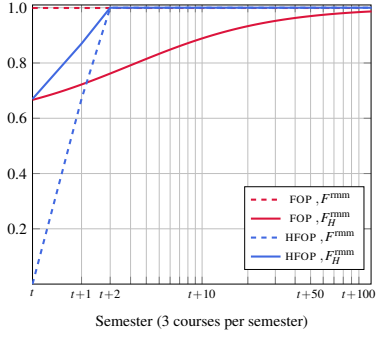
\begin{figure}[t]
    \centering
    \begin{tikzpicture}[scale=.65]
      \begin{axis}[
        xlabel=Semester (3 courses per semester),
        grid=both,
        xmin=1,  
        xmax=120,
        ymin=0.0,
        ymax=1.01,
        ytick={0.2, 0.4, 0.6, 0.8, 1.0},
        yticklabels={0.2, 0.4, 0.6, 0.8, 1.0},
        xtick={1, 2, 3, 4, 5, 6, 7,8,9,10,11, 21,31,41,51, 61,71,81,91,101},
        xticklabels={$t$,$t\!+\!1$,$t\!+\!2$,,,,,,,,$t\!+\!10$,,,,$t\!+\!50$,,,,,$t\!+\!100$},
        xticklabel style={font=\scriptsize},
        ymajorgrids=true,
        xmode=log,
        log ticks with fixed point,
        legend style={at={(.98, 0.35)},
                      anchor=north east,
                      font=\footnotesize,},
      ]

        \addplot[rred, dashed, very thick] coordinates {
            (1.0,0.999) (101,0.999)
        };  
        \addplot[
          domain=1:121, 
          samples=100,
          color=rred,
          very thick
        ] {1 - 5/(3 * (x - 1) + 15)};

        \addplot[rblue, dashed, very thick] coordinates {
            (1.0,0.0) (2,0.67) (3,1) (4,1) (5,1) (121,1)
        }; 

        \addplot[rblue, very thick] coordinates {
            (1.0,0.67) (2,0.87) (3,1) (4,1) (5,1) (121,1)
        }; 

        \legend{
        {$\fop\ , F^\text{rmm}$}, 
        {$\fop\ , F_H^\text{rmm}$}, 
        {$\hfop\ , F^\text{rmm}$}, 
        {$\hfop\ , F_H^\text{rmm}$}
        }
      \end{axis}
    \end{tikzpicture}
    \caption{Compares the fairness of the solutions computed using a \fop (red) and \hfop (blue) in a simple scenario with no quality metric $Q$.
    Each solution is evaluated with both the \emph{relative max-min} fairness metric $F^\text{rmm}$, and its historical variant $F_H^\text{rmm}$, shown respectively in dashed and solid lines. The $x$ axis is in log scale.}
    \label{fig:fop vs hfop}
    \vspace{6mm}
\end{figure}

More generally, it is interesting to compare solutions computed by \fop \emph{vs.} \hfop, and the respective fairness metric $F^\text{rmm}$ and historical fairness metric $F_H^\text{rmm}$.
Figure~\ref{fig:fop vs hfop} depicts these metrics under a simple course allocation scenario where we assume there exists no quality metric $Q$ and 3 courses per semester.
As expected, we observe that the fairness metric $F^{\text{rmm}}$ of the solutions computed by \fop is always maximized. In this case, \fop always returns $x_{(1.5,1.5)}$.
\hfop, on the other hand, starts by computing unfair solutions $x_{(0,3)}$ and $x_{(0.5,2.5)}$ at time steps $t$ and $t+1$. From $t+2$ onward, \hfop returns the fair solution $x_{(1.5,1.5)}$.
These different choices for the solutions have a significant impact on the way the historical unfairness is compensated.
Whereas \hfop maximizes the historical fairness in two time steps, we observe that after 10 semesters (or 30 courses) \fop only reaches a value of 0.9.
As anticipated in the end of the previous section, we conclude that \fop takes a long time to compensate existing historical unfairness.

\subsection{DHFOP: Incorporating Discounted Historical Fairness}\label{sec:dhfop}

We observed in Figure~\ref{fig:fop vs hfop} how slowly the historical unfairness would be compensated when following  the solutions produced by \fop. In fact, it turns out it would never be fully compensated -- since \fop always computes perfectly balanced schedules, the lecturing load gap of $5$ would remain unchanged. This behaviour may not fit many domains. It may become especially problematic when considering scenarios with long histories of unfair solutions.

A historical fairness metric $F_H$ allows the specification of an optimization problem \hfop that reasons over remnant historical unfairness.
We observed this may lead to solutions that seem unfair at time step $t$ when only considering the current time step (i.e., according to $F$).

In practice, it makes sense to consider a ``forgetting rate phenomenon'', where we attribute more importance to recent events than those in a distant past.
However, $F_H(x)$ puts equal importance to the fairness of solution $x$ at time step $t$ and all past solutions.
In order to model the importance of recent events we propose the discounted historical fairness metric $F_{H,\gamma}$, which discounts past unfairness with a forgetting discount factor $\gamma$.

\begin{definition}\label{HDFOP}
    A \textbf{Discounted Historical Fair Optimization Problem ($\mbox{DHFOP}$)} is a tuple $\langle Q, F_{H,\gamma}, H, \mathcal{X}, \mathcal{C},\beta\rangle$, where $F_{H,\gamma}$ is a historical fairness metric that reasons over a history of previous solutions $H = (x_{t-T}, \dots, x_{t-1})$ with the discount factor $\gamma$. The remaining elements follow the \hfop.
\end{definition}
Reasoning over a discounted historical fairness metric $F_{H,\gamma}$ allows us to control the importance of the unfairness of past solutions relative to more recent ones. Formally, the optimization problem is:
\begin{equation}
    \begin{array}{ll@{}ll}
    \underset{x\in\mathcal{X}}{\max} & Q(x) + \beta F_{H,\gamma}(x) & \\
    \ \ \text{s.t.} & \mathcal{C}(x)
\end{array}
\end{equation}

\noindent There are now two hyper-parameters.
The discount factor $\gamma$, which sets the importance of the fairness of past solutions,
and the parameter $\beta$, which controls the quality/fairness trade-off.

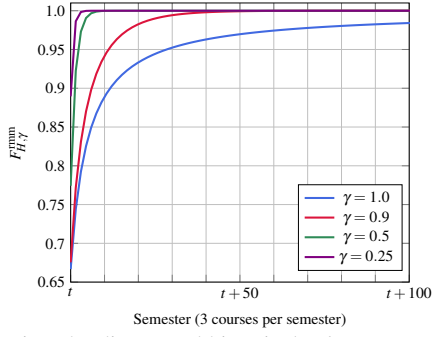
\begin{figure}[t]
    \centering
    \begin{tikzpicture}[scale=.65]
    \begin{axis}[
        xlabel=Semester (3 courses per semester),
        ylabel=$F_{H,\gamma}^\text{rmm}$,
        grid=both,
        xmin=0,
        xmax=100,
        ymin=0.65,
        ymax=1.01,
        ytick={0.65, 0.7, 0.75, 0.8, 0.85, 0.9, 0.95, 1.0},
        yticklabels={0.65, 0.7, 0.75, 0.8, 0.85, 0.9, 0.95, 1.0},
        xtick={0, 10, 20, 30, 40, 50, 60, 70, 80, 90, 100, 110, 120, 130, 140, 150},
        xticklabels={$t$,,,,,$t + 50$,,,,,$t+100$,,,,,$t+150$},
        ymajorgrids=true,
        log ticks with fixed point,
        legend pos=north west,
        ylabel style={yshift=-5pt},
        legend style={at={(.98, 0.35)},
                      anchor=north east},
    ]

    \addplot[
        domain=0:150,
        samples=100,
        color=rblue,
        very thick
    ] {1 - 5/(3 * x + 15)};

    \def\cgamma{0.9} 
    \addplot[
        domain=0:150,
        samples=100,
        color=rred,
        very thick
    ] {1 - (\cgamma^4 * 1 + \cgamma^2 * 3 + \cgamma) * \cgamma^x / (3 * (1 - \cgamma^(x + 5)) / (1 - \cgamma))};

    \def\cgamma{0.5} 
    \addplot[
        domain=0:150,
        samples=100,
        color=rgreen,
        very thick
    ] {1 - (\cgamma^4 * 1 + \cgamma^2 * 3 + \cgamma) * \cgamma^x / (3 * (1 - \cgamma^(x + 5)) / (1 - \cgamma))};

    \def\cgamma{0.25} 
    \addplot[
        domain=0:150,
        samples=100,
        color=violet,
        very thick
    ] {1 - (\cgamma^4 * 1 + \cgamma^2 * 3 + \cgamma) * \cgamma^x / (3 * (1 - \cgamma^(x + 5)) / (1 - \cgamma))};

      \legend{$\gamma=1.0$, $\gamma=0.9$, $\gamma=0.5$, $\gamma = 0.25$}
      \end{axis}
    \end{tikzpicture}
    \caption{Depicts the discounted historical \emph{relative max-min} fairness metric $F_{H,\gamma}^\text{rmm}$ in the setting where from time step $t=0$ onwards we schedule perfectly balanced loads. The smaller the discount factor $\gamma$, the faster the historical fairness is compensated.}
    \label{fig:compensating fairness debt gamma}
    \vspace{6mm}
\end{figure}

Revisiting once more our running example of course assignment and the max-min fairness metric, we could define its discounted historical variant as follows:
\begin{equation*}
    F_{H,\gamma}^\text{rmm}(x) = 1 - \frac{\max_{i} S_{\gamma, i}((H, x)) - \min_j S_{\gamma,j}((H, x))}{S_\gamma((H, x))},
\end{equation*}
where $S_{\gamma, i}((H, x)) = \sum_{\Delta=0}^{T} \gamma^{\Delta} S_i(x_{t - \Delta})$ is the discounted number of courses lectured by $l_i$ over all solutions in $(H, x)$ and similarly, $S_\gamma((H, x))$ is the discounted total number of courses lectured:
$$
S_\gamma((H,x)) = \sum_{\Delta=0}^{T} \gamma^{\Delta} S(x_{t - \Delta})
$$
We now analyze the behavior of $F_{H,\gamma}^\text{rmm}$ for different values of $\gamma$.
We build upon our course assignment example introduced in the previous section, assuming that at time step $t$ and onward we accept a solution that assigns an equal load to each lecturer (the solution that the \fop would compute).
Figure~\ref{fig:compensating fairness debt gamma} depicts $F_{H,\gamma}^\text{rmm}$ for different values of $\gamma$.
We observe that smaller values of $\gamma$ lead to a faster compensation of historical fairness.
For example, for $\gamma$ values of $0.25$, $0.5$, and $0.9$, it takes, $2$, $5$, and $26$ semesters, respectively, for $F_{H,\gamma}^\text{rmm}$ to reach a value of 0.99.

\subsection{MSDHFOP: Historically Fair Planning with Future Forecasts}\label{sec:mshfop}

All problems introduced so far are of a single-shot nature, where the solver is assumed to make a decision for the current time step $t$.
However, single-shot decisions can often result in sub-optimal solutions in complex domains with extended horizons, such as \textit{planning problems}~\cite{ghallab2016automated}.
Reasoning over multiple time steps into the future can allow for more effective solutions given knowledge or predictions about future events.

In the setting of fairness, reasoning over multiple steps into the future may allow for interesting solutions.
For example, due to future constraints, a solution that is fair over a given horizon may require initial solutions that seem unfair when analyzed independently.
In order to account for both existing historical unfairness and a planning horizon into the future, we let $F_{H,\gamma, \tau}(x_t,\dots,x_{t+T_F})$ denote a fairness metric that considers both a history of solutions $H$ and a sequence of $T_F$ future problems $(x_t, \dots, x_{t+T_F})$, with the past and the future being discounted according to $\gamma$ and $\tau$.

\begin{definition}\label{eq:MSHFOP}
    A \textbf{Multi Step Historical Fair Optimization Problem ($\mbox{MSDHFOP}$)} is a tuple $\langle Q, F_{H,\gamma,\tau}, H, \mathcal{X}, \mathcal{C},\beta, \gamma, \tau\rangle$, where $F_{H,\gamma,\tau}$ is a historical fairness metric that reasons over a history of previous solutions $H = (x_{t-T_H}, \dots, x_{t-1})$ and a sequence of future solutions $(x_t, \dots, x_{t + T_F})$. $\gamma$ and $\tau$ are discount factors. The remaining elements follow the \dhfop.
\end{definition}
We formulate an \msdhfop as:
\begin{equation}
    \begin{array}{ll@{}ll}
    \underset{x_0, \dots, x_{T_F}}{\max} & \!\sum_{t=0}^T\!\tau^t Q(x_t) + \beta F_{H,\gamma,\tau}(x_0, \dots, x_{T_F})  & \\
    \quad \  \text{s.t.} & \mathcal{C}(x)
\end{array}
\end{equation}
The first term computes the discounted sum of quality of the planned solutions.
The second term computes the multi-step historical fairness metric.
As before, the discount factor $\gamma$ sets the importance of past solutions relative to more recent ones in the computation of fairness.
Similarly, the discount factor $\tau$ discounts future solutions relative to the previous one, impacting both fairness and solution quality.
Whereas $\gamma$ seeks to model the ``recency effect'' from a fairness perspective (i.e., we tend to attribute more importance to recent events than those in a distant past), $\tau$ seeks to model uncertainty in planning into the future (i.e., it is easier to predict states closer in time than those in a distant future).

We can again build upon the discounted historical relative max-min fairness metric, and introduce a variant that also reasons over the next $T_F$ solutions $\boldsymbol{x}_{t:T_F} = (x_t, \dots, x_{t+T_F})$, where we define 
\begin{multline*}
    F_{H,\gamma,\tau}^\text{rmm}(\boldsymbol{x}_{t:T_F}) \\= 1 - \frac{\max_{i} S_{\gamma, \tau, i}((H, \boldsymbol{x}_{t:T_F})) - \min_j S_{\gamma,j}((H, \boldsymbol{x}_{t:T_F}))}{S_\gamma((H, \boldsymbol{x}_{t:T_F})},
\end{multline*}
where
\begin{equation*}
 S_{\gamma, \tau, i}((H, \boldsymbol{x}_{t:T_F})) = 
\sum_{\Delta=1}^{T_H} \gamma^{\Delta} S_i(x_{t - \Delta})
+
\sum_{\Delta=0}^{T_F} \tau^{\Delta} S_i(x_{t + \Delta})
\end{equation*}
is the discounted number of courses lectured by $l_i$, over all solutions in history $H$ and future planned solutions $\boldsymbol{x}_{t:T_F}$. Similarly, $S_{\gamma,\tau}((H, \boldsymbol{x}_{t:T_F}))$ is the discounted total number of courses lectured.

\begin{table}
  \vspace{3mm}
    \centering
        \caption{Builds upon the results in Table~\ref{tab:compare F and F_H}, reporting $F_{H,\gamma,\tau}$ for different solutions, assuming a history as depicted in Figure~\ref{subfig:unfair previous plans}, and $\gamma=\tau=1$. From Table~\ref{tab:compare F and F_H} we observed that a single-shot decision based on $F_H$ would pick solution $x_{(0,3)}$. However, if we are now aware there exists a constraint preventing $l_1$ from teaching any course at time step $t+1$, then the best decision is to first choose solution $x_{(0.5,2.5)}$.}
    \label{tab:ms single vs two}
    \small
    \begin{tabular}{@{}ccc@{}}
        $x_t$            & $x_{t+1}$      &   $F_{H,\gamma,\tau}((x_t, x_{t+1}))$ \\ \toprule
        $x_{(0,3)}$      & $x_{(0,3)}$   & 0.94 \\
        $x_{(0.5,2.5)}$  & $x_{(0,3)}$   & $\boldsymbol{1.00}$       \\
        $x_{(1.5,1.5)}$  & $x_{(0,3)}$   & 0.88               \\ \bottomrule
    \end{tabular}

\end{table}
Table~\ref{tab:ms single vs two} depicts an example where planning multiple steps into the future can lead to better solutions.
This example builds upon our analysis of Table~\ref{tab:compare F and F_H} from which we concluded the optimal solution according to $F_H$ is to assign at time step $t$ all the lecturing load to $l_2$.
However, suppose now we are allowed to plan over a horizon $T=2$ into the future, and that we are aware of a constraint preventing $l_1$ from lecturing any courses in the second semester $t+1$.
From Table~\ref{tab:ms single vs two} we conclude the best sequence of actions is actually $(x_{(0.5,2.5)}, x_{(0,3)})$.
Following \hfop instead would yield to the less rewarding solution $(x_{(0,3)}, x_{(0,3)})$.

\section{Experimental Evaluation}
\label{sec:experimental evaluation}
\subsection{Setup}

We evaluate our formulations across multiple domains using different fairness metrics.
We start with a technical description of each domain, introducing the decision variables, and the quality and fairness metrics to be used.
The machine used to run experiments is an Intel(R) Xeon(R) CPU E3-1585L v5 @ 3.00GHz  with 64GB of RAM.

\subsubsection{Course Assignment Problem (CAP)}
This is the domain that has been used throughout the paper, where a set of lecturers $\mathcal{L}$ is to be assigned to a set of courses $\mathcal{C}$.
When dealing with multi-step decision making settings, we may denote the set of courses at time step $t$ as $\mathcal{C}_t$.
The expertise of lecturer $l$ in course $c$ is measured by $S : \mathcal{L} \times \mathcal{C} \to \mathbb{R}$, and higher values correspond to higher expertise.
Decision variable $x_{l,c} \in \{0,  0.5, 1\}$ indicate the load of lecturer $l$ in teaching course $c$---
a lecturer may not lecture the course at all, or lecture either half a course or the full course.

We consider a quality metric
$Q = \frac{1}{Q_\text{max}}\sum_{c \in \mathcal{C}_t} \sum_{l \in \mathcal{L}} x_{l,c} \ S(l, c)$, which rewards course assignments with skilled lecturers. $Q_\text{max}$ is a normalization constant, denoting the maximum sum of expertise possible---this ensures $Q$ is bounded between $0$ and $1$.
In order to display the generality of our formulation, throughout the experimental evaluation with this domain we may use different fairness metrics. 

\subsubsection{Vehicle Routing Problem (VRP)}
  In this domain, given a set of vehicles $V$, a set of points $P$ that must all be traveled to 
  exactly once, a depot $r \in P$ the vehicles must leave from and return to, and distances between all points 
  $D : P \times P \to \R_+$, determine a 
  route for each vehicle that minimizes the total distance traveled.
  We consider a standard integer program to model the \op where quality $Q(x)$
  is the total distance traveled (see 
  Supplementary Materials~\ref{sec:ip-definitions-app} for a full definition).
  To model fairness, for a given solution $x$ to the integer program, let $U_v(x)$ be the total distance vehicle $v$
  travels under $x$ and define $F(x):= \max_{v \in A} U_v(x) - \min_{w\in A} U_w(x)$. This notion of fairness is similar to 
  proportional equality and is used in~\cite{jozefowiez2009evolutionary} in a multi-objective version of VRP.

\subsubsection{Task Allocation Problem (TAP)}
  In this domain, given a set of agents $A$, a set of tasks $T$, and a cost 
  associated with each agent for each task $C : A \times T \to \R_+$, find an assignment of tasks to agents such that the
  sum of costs is minimized. We consider a standard integer program to model the \op 
  where quality $Q(x)$ is the sum of costs (see Supplementary 
  Materials~\ref{sec:ip-definitions-app} for a full definition).
  \iflongversion
  The fairness metric we consider is the classic minimax notion of fairness (see~\cite{xinying2023guide} and 
  references therein), where for a given solution 
  $x$ to the integer program, let $U_a(x)$ be the total cost agent $a$ incurs under $x$ and define 
  $F(x) := \max_{a \in A} U_a(x)$.
  \else
  The fairness metric we consider is the classic minimax notion of fairness (see~\cite{xinying2023guide} and 
  references therein).
  \fi

\subsubsection{Nurse Scheduling Problem (NSP)}
We consider a version of this classical problem in operations research. 
In the Supplementary Materials~\ref{sec:exp-dist-fair-hist}, we first formally 
define the problem and then we show the impact of 
different histories on the NSP, in particular showcasing the impact of the
discount factor in \dhfop.

\subsection{Quality vs. Fairness}

\begin{figure}[t]
    \centering
    \begin{tikzpicture}[scale=0.5]
    \begin{axis}[
        xmin=0, xmax=9,
        ymin=-6.5, ymax=1.01,
        xtick={0, 1, 2, 3, 4, 5, 6, 7, 8, 9},
        xticklabels={$t$,,$t\!+\!2$,,$t\!+\!4$,,$t\!+\!6$,,$t\!+\!8$},
        grid=both,
        xlabel={Semester},
        xlabel style={yshift=5pt},
        yticklabel style={font=\small},
        ylabel style={yshift=-15pt},
        legend style={at={(.98, 0.7)},
                      anchor=north east},
        title={$\hfop \ \beta=0.125$},
        title style={yshift=-5pt},
        ]
        \addplot[rgreen,very thick,mark=*] coordinates {
            (0, 1.0) (1, 0.75) (2, 0.875) (3, 0.875) (4, 0.875) (5, 0.4375) (6, 0.4375) (7, 0.875) (8, 0.4375) (9, 0.4375)
        };
    
        \addplot[rred,very thick,mark=*] coordinates {
             (0, -1.0) (1, -1.0) (2, -0.25) (3, -0.25) (4, -0.25) (5, -0.0625) (6, -0.0625) (7, -0.25) (8, -0.0625) (9, -0.0625)
        };

        \addplot[rblue,very thick,mark=*] coordinates {
            (0, -1.0) (1, -1.0) (2, -2.25) (3, -4.0) (4, -6.25) (5, -5.0625) (6, -4.0) (7, -6.25) (8, -5.0625) (9, -4.0)
        };

        \legend{$Q$, $F$, $F_H$}
    \end{axis}
    \end{tikzpicture}
    \begin{tikzpicture}[scale=0.5]
    \begin{axis}[
        xmin=0, xmax=9,
        ymin=-2.5, ymax=1.01,
        xtick={0, 1, 2, 3, 4, 5, 6, 7, 8, 9},
        xticklabels={$t$,,$t\!+\!2$,,$t\!+\!4$,,$t\!+\!6$,,$t\!+\!8$},
        yticklabel style={font=\small},
        grid=both,
        xlabel={Semester},
        xlabel style={yshift=5pt},
        legend style={at={(.98, 0.2)},
                      anchor=north east},
        title={$\hfop \ \beta=0.25$},
        title style={yshift=-5pt},
        ]
        \addplot[rgreen,very thick,mark=*] coordinates {
             (0, 0.875) (1, 0.875) (2, 0.875) (3, 0.4375) (4, 0.4375) (5, 0.875) (6, 0.4375) (7, 0.4375) (8, 0.875) (9, 0.4375)
        };
    
        \addplot[rred,very thick,mark=*] coordinates {
            (0, -0.25) (1, -0.25) (2, -0.25) (3, -0.0625) (4, -0.0625) (5, -0.25) (6, -0.0625) (7, -0.0625) (8, -0.25) (9, -0.0625)
        };

        \addplot[rblue,very thick,mark=*] coordinates {
            (0, -0.25) (1, -1.0) (2, -2.25) (3, -1.5625) (4, -1.0) (5, -2.25) (6, -1.5625) (7, -1.0) (8, -2.25) (9, -1.5625)
        };

    \end{axis}
    \end{tikzpicture}
    \\
    \begin{tikzpicture}[scale=0.5]
    \begin{axis}[
        xmin=0, xmax=9,
        ymin=-.3, ymax=1.01,
        xtick={0, 1, 2, 3, 4, 5, 6, 7, 8, 9},
        xticklabels={$t$,,$t\!+\!2$,,$t\!+\!4$,,$t\!+\!6$,,$t\!+\!8$},
        grid=both,
        xlabel={Semester},
        xlabel style={yshift=5pt},
        yticklabel style={font=\small},
        legend style={at={(.98, 0.2)},
                      anchor=north east},
        title={$\hfop \ \beta=0.75$},
        title style={yshift=-5pt},
        ]
        \addplot[rgreen,very thick,mark=*] coordinates {
              (0, 0.875) (1, 0.4375) (2, 0.6875) (3, 0.625) (4, 0.4375) (5, 0.6875) (6, 0.625) (7, 0.4375) (8, 0.6875) (9, 0.625)
        };
    
        \addplot[rred,very thick,mark=*] coordinates {
             (0, -0.25) (1, -0.0625) (2, -0.0625) (3, -0.0625) (4, -0.0625) (5, -0.0625) (6, -0.0625) (7, -0.0625) (8, -0.0625) (9, -0.0625)
        };

        \addplot[rblue,very thick,mark=*] coordinates {
             (0, -0.25) (1, -0.0625) (2, -0.25) (3, -0.25) (4, -0.0625) (5, -0.25) (6, -0.25) (7, -0.0625) (8, -0.25) (9, -0.25)
        };

    \end{axis}
    \end{tikzpicture}
    \begin{tikzpicture}[scale=0.5]
    \begin{axis}[
        xmin=0, xmax=9,
        ymin=-.3, ymax=1.01,
        xtick={0, 1, 2, 3, 4, 5, 6, 7, 8, 9},
        xticklabels={$t$,,$t\!+\!2$,,$t\!+\!4$,,$t\!+\!6$,,$t\!+\!8$},
        grid=both,
        xlabel={Semester},
        xlabel style={yshift=5pt},
        yticklabel style={font=\small},
        legend style={at={(.98, 0.2)},
                      anchor=north east},
        title={$\hfop \ \beta=2$},
        title style={yshift=-5pt},
        ]
        \addplot[rgreen,very thick,mark=*] coordinates {
               (0, 0.6875) (1, 0.625) (2, 0.4375) (3, 0.6875) (4, 0.625) (5, 0.4375) (6, 0.6875) (7, 0.625) (8, 0.4375) (9, 0.6875)
        };
    
        \addplot[rred,very thick,mark=*] coordinates {
             (0, -0.0625) (1, -0.0625) (2, -0.0625) (3, -0.0625) (4, -0.0625) (5, -0.0625) (6, -0.0625) (7, -0.0625) (8, -0.0625) (9, -0.0625)
        };

        \addplot[rblue,very thick,mark=*] coordinates {
             (0, -0.0625) (1, -0.0625) (2, -0.0) (3, -0.0625) (4, -0.0625) (5, -0.0) (6, -0.0625) (7, -0.0625) (8, -0.0) (9, -0.0625)
        };

    \end{axis}
    \end{tikzpicture}
    \caption{The quality $Q$, fairness $F^\text{qmmg}$, and historical fairness $F_H^\text{qmmg}$ of the solutions computed by $\hfop$ under different values of $\beta$, in the course assignment domain.}
    \label{fig:quality vs fairness}
    \vspace{6mm}
\end{figure}
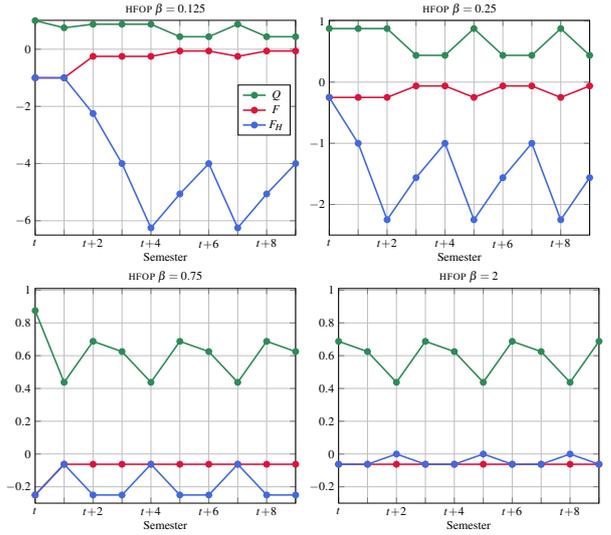

We start our experimental evaluation with an example depicting the quality \emph{vs.} fairness trade-off, and the impact of the parameter $\beta$ therein.
Let's consider an instance of \textsc{CAP} with 3 lecturers, $l_1, l_2, l_3$ and 2 courses $c_1, c_2$.
Across all 3 courses, $l_1$ has high skills $(S = 2)$, $l_2$ has medium skills $(S = 1.5)$, and $l_3$ has low skills $(S = 0)$.
We use a historical \emph{quadratic max-min gap} fairness metric 
\[
F^\text{qmmg}_{H}(x) = -\left( \frac{1}{2} \ ( \max_i S_{i}( (H,x) ) - \min_j S_{j}((H,x)) ) \right)^2,
\]
where $S_{\gamma, i}$ is defined as previously.
The original fairness metric that disregards $H$ follows naturally.
$F^\text{qmmg}$ has range $[-\infty, 0]$ and,  when compared to $F^\text{rmm}$, should allow for heavier penalization of solutions that increase the lecturing load, due to the quadratic term and the lack of normalization.

We analyzed the solutions computed by \hfop under different values of $\beta$, for 10 consecutive semesters, starting with no previous history.
Figure~\ref{fig:quality vs fairness} compares the quality $Q$, fairness $F^\text{qmmg}$, and historical fairness $F_H^\text{qmmg}$ of the solutions computed by \hfop under different values of $\beta$.
Table~\ref{tab:quality vs fairness} provides a summary of the results for the different metrics.
We observe that, as $\beta$ increases, \hfop computes solutions with lower quality $Q$, but higher fairness $F$ and historical fairness $F_H$.
This follows our expectation, since $\beta$ is the parameter setting the quality \emph{vs.} fairness trade-off.
From the figure we also observe that \hfop converges to a pattern of first selecting higher-quality/lower-fairness solutions, and once the historical fairness reaches a certain (low) level, starts selecting lower-quality/higher-fairness solutions to compensate for it.
This pattern is even more noticeable for lower values of $\beta$, where at earlier time steps the solutions produced tend to be characterized by high quality and low fairness.

\begin{table*}[!ht]
\centering
\caption{Summary of the results on the comparison of the quality $Q$, fairness $F^\text{qmmg}$, and historical fairness $F_H^\text{qmmg}$ of the solutions computed by $\hfop$ under different values of $\beta$. }
\label{tab:quality vs fairness}
\renewcommand{\arraystretch}{0.75}
\small
\begin{tabular}{@{}ccccccccccc@{}}

\toprule
                      &      & \multicolumn{3}{c}{$Q$}                 & \multicolumn{3}{c}{$F^\text{qmmg}$}                  & \multicolumn{3}{c}{$F_H^\text{qmmg}$}                 \\ \midrule
                      &      & max & min & $\mu \pm \sigma$ & max & min & $\mu \pm \sigma$ & max & min & $\mu \pm \sigma$ \\
\multicolumn{2}{c}{\op} &        1.0 & 1.0 & $1.0 \pm 0$ &  $-1.0$ & $-1.0$ & $-1.0 \pm 0.0$     &  $-1.0$ & $-100.0$ & $-38.5 \pm 2.42$                          \\ \midrule
\multirow{5}{*}{\rotatebox[origin=c]{90}{\hfop}} & $\beta$    &     &     &     &     &     &     &     &  \\
     & $0.125$ & $1.0$  & $0.44$ & $0.7 \pm 0.22$ & $-0.06$ & $-1.0$ & $-0.33 \pm 0.35$ &  $-1.0$ & $-6.25$ & $-3.89 \pm 1.83$ \\
     & $0.25$   &  $0.88$  &  $0.44$ &  $0.66 \pm 0.22$  &   $-0.06$  &   $-0.25$  &  $-0.16 \pm 0.09$  &  $-0.25$   &   $-2.25$  &  $-1.47 \pm 0.63$     \\
     & $0.75$   &  $0.88$   &   $0.44$  &   $0.61 \pm 0.13$   &  $-0.06$   &   $ -0.25$  &  $-0.08 \pm 0.06$  &  $-0.06$   &   $-0.25$  &   $-0.19 \pm 0.09$   \\
     & $2$    &  $0.69$   &  $ 0.44$   &    $0.59 \pm 0.11$   &   $-0.06$  &   $-0.06$  &   $-0.06 \pm 0.0 $   &    $-0$ &   $-0.06$  &    $-0.04 \pm 0.03$   \\ \bottomrule
\end{tabular}
\vspace{2mm}
\end{table*}

\subsection{Planning with Future Forecasts}

We now evaluate the benefits from a fairness perspective of \msdhfop reasoning over multiple steps into the future from a fairness perspective.
Consider a simplified instance of CAP with two lecturers $l_1$ and $l_2$, and two courses $c_1$ and $c_2$.
Across all courses, $l_1$ has high skills ($S = 2$) and $l_2$ has medium skills ($S = 1$).
To showcase the flexibility of our approach to fairness metrics, we now consider
another version of the \emph{maximin} fairness metric where utility $U_i$ measures the number of courses taught by lecturer $l_i$:
\begin{equation*}
    F^\text{mm}(x) = \frac{\min_{i} U_i(x)}{\max_{j} U_j(x)}.
\end{equation*}
Assuming no historical solutions, the scheduler is now to plan the course assignments for the next $T=4$ semesters.
There exists a known constraint about the future---$l_1$ will not  be able to take any lecturing load on semesters $t\!+\!2$ and $t\!+\!3$ due to a sabbatical leave.

Table~\ref{tab:planning future} depicts the solutions computed by \hfop and \msdhfop, in a setting with $\beta=2$ and discount factors $\gamma=1, \tau=1$.
Since $\hfop$ plans a single step at a time, it is not able to take advantage of the information on $l_1$'s future constraints.
As a result, it schedules the perfectly balanced solution in the two initial steps, and is then forced to schedule the last two time steps as $x_{(0,2)}$.
This results in a sequence of solutions leading to an overall lower quality (10 \emph{vs.} 12) and fairness ($0.33$ \emph{vs.} $1$).

\addtolength{\tabcolsep}{-2pt}
\vspace{3mm}
\begin{table}
    \centering
        \caption{Comparison between \fop and \msdhfop, showing the benefits of reasoning over multiple steps into the future.
        Since \hfop performs single-shot decisions, it ends up ignoring the known constraint that $l_1$ will not lecture any course in time steps $t\!+\!2$ and $t\!+\!3$.}
    \label{tab:planning future}
    \footnotesize
    \begin{tabular}{@{}ccccccc@{}}
               & $x_t$        & $x_{t\!+\!1}$    & $x_{t\!+\!2}$    & $x_{t\!+\!3}$   & $\sum_t Q(x_t) $ & $F_{H,\gamma,\tau}(\boldsymbol{x})$ \\ \toprule
      \hfop    & $x_{(1,1)}$  & $x_{(1,1)}$  & $x_{(0,2)}$  & $x_{(0,2)}$ & 10               & 0.33 \\
      \msdhfop & $x_{(2,0)}$  & $x_{(2,0)}$  & $x_{(0,2)}$  & $x_{(0,2)}$ & $\boldsymbol{12}$               & $\boldsymbol{1.0}$       \\ \bottomrule
    \end{tabular}

\end{table}
\addtolength{\tabcolsep}{2pt}

\subsection{Increasing Complexity and Benchmarking}
  We now examine a more complex problem and show the impact of considering fairness
  on running time. 
  We first introduce a method for generating random instances and a history of past solutions for VRP. Consider a square 
  integer grid of a fixed size. 
  We deterministically place the depot at the center of the grid and, given a fixed number of points $n$,
  choose $n$ of the grid points uniformly at random (not including the depot). For generating history, we generate
  random instances and solve the problem optimally on these random instances.

  For our experiments in this section and the following, we implemented the integer program with the 
  corresponding fairness constraints
  using the PuLP Python library~\cite{mitchell2011pulp} and used the CBC solver~\cite{forrest2005cbc} 
  with a standard linearization of $F(x)$ (see, e.g.,~\cite{xinying2023guide}). All times measured are 
  wall-clock times for the combined model-building and solving times.

  In our experiments, we consider 4 vehicles $V = \{V_1,V_2,V_3,V_4\}$ and 12 locations. We generate a history of 5 steps 
  and a single random instance. For each historical instance, we assume $V_1$ always had the shortest route, 
  $V_2$ the second shortest, and similarly for $V_3$ and $V_4$. Table~\ref{tab:vrp-history} shows the total distance 
  traveled for each vehicle. We compare the solutions of \op, \fop, and \hfop. We set $\beta = 10$ for all experiments. 
  Table~\ref{tab:vrp-distances} shows the results for each of the 4 vehicles.
  
  In \fop, the notion of fairness considered should encourage solutions where all distances traveled are similar. This, of course, should come at the 
  expense of increasing the overall distance traveled. We see this exact scenario play out when comparing \op and 
  \fop. The distances in the solution for \op are not uniform but attain a total distance of $74.4$ and the distances
  in the solution for \fop are all similar but the total distance traveled is $112.1$. 
  When comparing \op and \fop to \hfop, we expect that \hfop should account for the historical unfairness received
  by vehicle $V_4$. 
  In fact, we expect and see in the results that the solution to \hfop should give the shortest
  routes (in order) to $V_4$, $V_3$, $V_2$ and $V_1$.  
  In terms of total time,  \fop and \hfop require
  roughly 3 times as long to run, thus showing that the cost of incorporating fairness in our framework is not 
  computationally prohibitive for VRP.
  \begin{table}
    \vspace{3mm}
    \centering    
    \caption{Results for VRP experiments.}
    \vspace{2mm}
    \begin{subtable}{\linewidth}
      \small
      \centering
      \begin{tabular}{cccc}
        $V_1$ & $V_2$ & $V_3$ & $V_4$\\\toprule
        37.1 & 44.9 & 154.4 & 202.6 \\\bottomrule
      \end{tabular}
      \vspace{4mm}
      \caption{Total historical distance traveled.}
      \label{tab:vrp-history}
      
    \end{subtable} 

    \vspace{1mm}
    
    \begin{subtable}{\linewidth}
      \small
      \centering
      \begin{tabular}{cccccc}
        & $V_1$ & $V_2$ & $V_3$ & $V_4$ & time (s) \\ \toprule
        \op & 4.5 & 6.3 & 13.8 & 49.8 & 11.6 \\ \midrule
        \fop & 26.6 & 28.3 & 28.6 & 28.6 & 34.2 \\ \midrule
        \hfop & 74.5 & 65.3 & 6.32 & 4.47 & 33.9 \\ \bottomrule
      \end{tabular}
      \vspace{4mm}
    \caption{Distance traveled for all 4 vehicles on a single instance.}
    \label{tab:vrp-distances}
    \end{subtable}

  \end{table}

\subsection{Larger Scale Experimentation}

  In this section, we show that our framework can be applied on a larger scale than considered in the
  previous sections. 
  We introduce a method for generating random instances for TAP.
  We create instances with $|A| = |T| = 40$. For each agent $a \in A$, one task is chosen uniformly at random
  to have cost $5$, three tasks are chosen uniformly at random to have cost $20$, and the rest of the tasks have 
  cost $30$. We sometimes deterministically enforce that an agent $a$ does not have a task of cost $5$ and this 
  task is replaced with a cost $30$ task, in which case we say agent $a$ is \emph{constrained}. 
  
  We refer to the following setup as a single \emph{run} and we average our results over 10 runs. Sample $8$ 
  agents uniformly at random from $A$ and denote this subset as $C$. Produce 3 random instances according to 
  our random instance generation given above. Then produce 3 more random instances where all agents in 
  $C$ are constrained. These 6 instances are the future instances. To generate history, we run the \op
  on each of these 6 instances. Sort the agents according to total cost. In this order, the last $4$
  agents not in $C$ are assigned a historical cost of $180$, call these agents $W$. 
  Amongst the remaining agents, the first $24$ agents 
  are assigned a historical cost of $30$. The remaining $12$ agents are assigned a historical cost of $120$.
  We give more justification for our method of random instance generation and history generation in the Supplementary 
  Materials~\ref{sec:tap-instance-explanation}.
  All instances are run with $\beta = 10$.
  
We first evaluate the maximum cost assigned to any agent in \op and \fop. Table~\ref{tab:tap-worst-off} shows the number of times per run the maximum cost is $30$. We expect the number to be much larger in \op compared to \fop, which is confirmed in Table~\ref{tab:tap-worst-off}. This is at the expense of incurring a larger total cost, which is expected since \fop is also prioritizing minimizing the maximum cost and not just the total cost.

We next evaluate the impact of history in \op, \fop, and \hfop. Recall that for each run, $W$ is the set of agents who received the largest historical cost. \op and \fop do not consider history and therefore will not necessarily prioritize the agents in $W$. We see this exact behavior in Table~\ref{tab:tap-worst-off}. Furthermore, even as \op continues to not prioritize the agents in $W$, which is how $W$ is defined, the average cost of agents not in $W$ for \op is still less than for \fop or \hfop. This is expected, at least for \hfop, as \hfop prioritizes agents in $W$.

We now examine the outcomes of the constrained agents $C$, who are all constrained in the last 3 instances of each run. Table~\ref{tab:tap-constrained} shows that the constrained agents in \op, \fop, and \hfop all have a similar average over both the first and second 3 instances, which is expected as these agents should not necessarily receive special treatment in any of these frameworks. However, \msdhfop reasons about the future, and therefore we expect it to adjust for the fact that the agents in $C$ are constrained over the last 3 instances. We see this behavior in Table~\ref{tab:tap-constrained}. Note that we set $\gamma = \tau = 0.75$.
  
  We also report the running times in Table~\ref{tab:tap-constrained}. The running times
  of \hfop and \msdhfop are about the same as \op, especially considering
  \msdhfop runs all 6 future instances at once. \fop, however, has a large
  average running time, but the median running time is only 1.6 s---some instances 
  require large amounts of time but are not common. One hypothesis is that the
  solver we use takes a lot of time when trying to minimize the max cost when
  there are multiple agents that can achieve the max cost. 
  In \hfop on the
  first future instance, for example, the
  historical imbalance ensures that only the 4 agents in $W$ can achieve the max cost,
  which may reduce the set of candidate optimal solutions considerably.

  \begin{table}[t]
  \renewcommand{\arraystretch}{0.75}
  \small
  \centering
      \caption{Evaluating \op, \fop, \hfop for TAP experiments.
    (First column) Measures the number of times the maximum cost is $30$---the largest value is $6$ as there are $6$ instances per run.
    (Last two columns) For each run, we compute the total 
    cost for each agent in $W$ and average it by $|W| = 4$, similarly for $A \setminus W$.
    All values averaged over 10 runs.}
    \begin{tabular}{ccccc}
      & \makecell{max\\cost=30} & \makecell{avg sum\\of costs} 
        & \makecell{avg cost \\ of $W$} & \makecell{avg cost \\ of $A \setminus W$} \\ \toprule
      \op & $5.9 \pm 0.3$ & $470.8\pm 12.1$ & $97.6 \pm 7.8$ & $67.6 \pm 1.7$\\ \midrule
      \fop & $2.4\pm0.8$ & $484.0\pm13.2$ & $86.7 \pm 9.3$ & $71.0 \pm 2.2$\\ \midrule
      \hfop & $6.0\pm0.0$ & $478.2\pm12.5$ & $50.7 \pm 7.7$ & $74.0 \pm 1.9$ \\ \bottomrule
    \end{tabular}

    \label{tab:tap-worst-off}
  \end{table}

    \begin{table}[t]
    \renewcommand{\arraystretch}{0.75}
  \small
  \vspace{5mm}
  \centering
    \caption{Evaluating the impact of the constrained agents on the TAP experiments. For each run, we 
    compute the total cost for each agent in $C$ and average it by $|C| = 8$. We compare
    the results over the first 3 instances and last 3 instances of each run.
    The times for each run are averaged over the 6 instances. All values averaged over
    10 runs.}
    \begin{tabular}{cccc}
      & \makecell{avg cost of\\ $C$ (first 3)} & 
        \makecell{avg cost of\\ $C$ (last 3)} & time (s) \\ \toprule
      \op & $32.5 \pm 4.4$ & $67.8 \pm 1.7$ & $0.5 \pm 0.1$ \\ \midrule
      \fop & $32.3 \pm 3.2$ & $62.1 \pm 2.1$ & $54.4\pm42.5$ \\ \midrule
      \hfop & $36.6 \pm 4.5$ & $67.3 \pm 1.9$ & $0.6 \pm 0.1$ \\ \midrule
      \msdhfop & $23.3 \pm 3.7$ & $64 \pm 0.9$ & $4.2\pm1.2$ \\ \bottomrule
    \end{tabular}
    \label{tab:tap-constrained}
  \end{table}

\section{Related Work}
\label{sec:related work}

In recent times, a significant amount of research has been dedicated to fairness in AI, with a focus on predictive models and algorithmic fairness~\cite{kleinberg2018algorithmic}.
In machine learning, in particular, the topic of long-term fairness has been the subject of much attention~\cite{d2020fairness,ge2021towards,hu2022achieving}.
The long-term consideration of fairness is relevant to the \msdhfop formulation, where we consider both future and past history.

Other lines of research look into the connections of algorithmic fairness and ethical decision making in the context of sequential decision making and planning~\cite{nashed2023fairness}.
Nashed et al. explore how each of these settings has articulated its normative concerns, the viability of different techniques for these different settings, and how ideas from one may be useful for the other.

Motivated by computational resource allocation problems, there exists a vast literature on the topic of fairness in real-time scheduling. 
Examples include the fair scheduling of periodically arriving tasks with deadlines~\cite{baruah1995fairness}, or more generally, the problem of scheduling tasks to long lived processes while taking into account the benefit/cost to each process~\cite{ajtai1998fairness}.
While these works look at fairness from a temporal perspective---seeking to ensure a fair load to the different processes---they do not consider possible historical unfairness due to previous solutions and tend to focus on a specific fairness metric.

In the areas of decision-making and planning, several works have focused on different ways to mathematically formulate fairness metrics. 
Recent work surveys various schemes that have been proposed for formulating ethics-related criteria, including those that integrate efficiency and fairness concerns~\cite{xinying2023guide}.
They emphasize the challenges of having a single definition of fairness, as different definitions are appropriate for different contexts.
Additionally, different fairness models are grouped into clusters, each representing a different type of fairness principle,  to facilitate comparisons and help identify the most suitable model for practical applications.
While the fairness metrics introduced did not consider fairness from a temporal perspective where a history of past solutions exists, they can be adapted and used as part of all our formulations. 

There has also been a growing interest in fairness in multi-agent decision making, planning~\cite{pozanco2022fairness}, and reinforcement learning~\cite{jiang2019learning,grupen2022cooperative}.
Recent work focuses on fairness in long-term decision making problems, introducing a new voting formalism that takes the history of previous decisions into account~\cite{lackner2020perpetual}. 
While the concept of considering history is similar to the definition of \hfop, our formalism considers a centralized decision making process.

\section{Conclusion and Future Work}
\label{sec:conclusion}
In this work we took a new angle to considering fairness in decision making processes.
Building upon previous fairness formulations, we focused on how to reason about fairness from a temporal perspective, especially when there exists a history of past decisions that may have been potentially unfair.
In this setting, we proposed to reason over the concept of \textit{``temporal fairness''} in decision making processes.

Starting from a general decision making problem---\op---we incrementally built our approach to reason over temporal fairness, accounting for both past solutions and predictions about the future.
With the introduction of a fairness metric in the objective, \fop extends \op by reasoning over the quality/fairness trade-off of a solution.
To reason over historical unfairness, we propose \hfop, where the fairness metric takes into account a history of previous solutions.
A discounted version \dhfop is then proposed to allow us to model the importance of more recent events.
Finally, the \msdhfop formulation is extended to reason over both historical and future solutions.
In the experimental evaluation we assess our approach across different domains and show, in particular, how our approach is compatible with different fairness metrics.

As directions for future work, we envision exploring scenarios where different fairness metrics are used across time (in the past and future) and reasoning over multiple concurrent fairness metrics.

\section{Disclaimer}
This paper was prepared for informational purposes in part by
the Artificial Intelligence Research group of JPMorgan Chase \& Co. and its affiliates (``JP Morgan''),
and is not a product of the Research Department of JP Morgan.
JP Morgan makes no representation and warranty whatsoever and disclaims all liability,
for the completeness, accuracy or reliability of the information contained herein.
This document is not intended as investment research or investment advice, or a recommendation,
offer or solicitation for the purchase or sale of any security, financial instrument, financial product or service,
or to be used in any way for evaluating the merits of participating in any transaction,
and shall not constitute a solicitation under any jurisdiction or to any person,
if such solicitation under such jurisdiction or to such person would be unlawful.

\bibliography{refs}

\appendix
\section{Additional Experiments}

\subsection{Integer Program Definitions of VRP \& TAP}
\label{sec:ip-definitions-app}
  In this section, we formally define the standard integer programs used in the experiments
  section. 
  \subsubsection{VRP Integer Program}
    Recall that $P$ is the set of points, $V$ is the set of vehicles, $r \in P$ is the
    depot, and $D$ is the distance function between all points.
    For convenience, let $P^- := P \setminus \{r\}$
    and $\cS = \{S \subseteq P^- \mid 2 \le |S| \le n-2\}$. Note that the binary variable $x_{a,b,v}$ is $1$ if and only if
    vehicle $v$ is routed from point $a$ to point $b$.
    \[
    \begin{array}{rl}
        \text{min} & \sum_{a \in P} \sum_{b \in P} \sum_{v \in V} x_{a,b,v} D(a,b)\\
        \text{s.t.} & \sum_{a \in P} x_{a,b,v} = \sum_{a \in P} x_{b,a,v}, \forall b \in P, \forall v \in V\\
        & \sum_{a \in P \setminus \{b\}} \sum_{v \in V} x_{a,b,v} = 1, \forall b \in P^-\\
        & \sum_{a \in P^-} x_{r, a,v} = 1, \forall v \in V \\
        & \sum_{a \in S} \sum_{b \not\in S} \sum_{v \in V} x_{a,b,v} \ge 1, S \in \cS\\
        & x_{a,b,v} \in \{0,1\}, \forall a,b \in P, \forall v \in V
    \end{array}
    \]
  \subsubsection{TAP Integer Program}
    Recall that $A$ is the set of agents, $T$ is the set of tasks, and $C(a,t)$ is the
    cost agent $a$ incurs to perform task $t$.
    We assume that 
    $|A| = |T|$. The binary variable $x_{a,t}$ is $1$ if and only if task $t$ is assigned to agent $a$.
    \[
      \begin{array}{rl}
        \text{min} & \sum_{a \in A} \sum_{t \in T} x_{a,t}C(a,t)\\
        \text{s.t.} & \sum_{a \in A} x_{a,t} = 1, \forall t \in T \\
        & \sum_{t \in T} x_{a,t} = 1, \forall a \in A \\
        & x_{a,t} \in \{0,1\}, \forall a \in A, \forall t \in T
      \end{array}
    \]

\subsection{TAP Instance Generation and History}
\label{sec:tap-instance-explanation}
  In this section, we provide more of a justification for the random instance 
  and history generation for TAP we use in our experiments. 
  We can conceptualize these instances as workers being assigned tasks. The workers either take $5$, $20$, 
  or $30$ minutes to complete a task and sometimes a worker does not have a $5$-minute task in the current batch of tasks,
  maybe due to a lack of expertise.
  Minimizing the total sum of costs corresponds to minimizing the total person-hours required to complete all tasks,
  whereas minimizing the maximum cost corresponds to minimizing the amount of time any one person has to spend on a task.

  Regarding the history we construct, we want to discuss the agents $W$ who are 
  assigned the largest historical cost of $180$. The amount $180$ comes from the
  hypothetical scenario where for 6 instances straight, all agents in $W$ received
  a task of cost $30$. Further, we choose $W$ to be the agents with the largest total
  cost not in $C$ for two reasons: (1) we want to see the dynamics of the constrained
  agents in $C$ without giving them the most ``historical debt" and (2) \op and \fop 
  treat all agents identically, so it is conceivable that the agents who received
  the largest cost under \op could also have the largest historical debt.
  
  As for how we constructed the rest of the history, the 12 agents who were assigned 
  a historical cost of $120$ were given that value under the hypothetical scenario where
  they were assigned cost 20 tasks for 6 instances. Similarly for the 24 agents who
  were assigned a historical cost of $30$ - they would have received a cost 5 task
  for 6 instances straight. 

\subsection{Distribution of Fairness in History}
\label{sec:exp-dist-fair-hist}
\subsubsection{Nurses Scheduling Problem (NSP)}
In this domain, a set of nurses $\mathcal{N}$ is to be assigned to a set of morning/evening shifts $\mathcal{S}$ across 5 days of the week. We let $\mathcal{S}_m$ and $\mathcal{S}_e$ denote the morning and evening shifts, respectively.
The decision variable $x_{n,s} \in \{0, 1\}$ indicates whether a nurse $n$ is assigned to shift $s$.
The seniority of the nurses is represented by $S : \mathcal{N} \to \mathbb{R}$ (higher means more senior).
The nurses may have preferences over some shifts, and this is represented through a utility function $\mathcal{U} : \mathcal{N} \times \mathcal{S} \to \{ 0, 1, 2, 3 \}$.

The quality metric considered looks to reward assignments of senior nurses to the evening shifts, as they tend to be the most problematic ones: $Q = \frac{1}{Q_\text{max}} \sum_{s \in \mathcal{S}_e} \sum_{n \in \mathcal{N}} x_{n,s} \ S(n)$. $Q_\text{max}$ is again a normalization constant. For the fairness metric, we use the same version of the \textit{maximin} metric as we used for the course assignment domain:
\begin{equation*}
    F^\text{mm}(x) = \frac{\min_{i} U_i(x)}{\max_{j} U_j(x)},
\end{equation*}
where $\min_{i} U_i(x)$ and $\max_{j} U_j(x)$ denote the minimum and maximum utility among all nurses, respectively.
Note that $F^\text{mm}$ is bounded in $[0,1]$.

\subsubsection{Experiments for NSP}
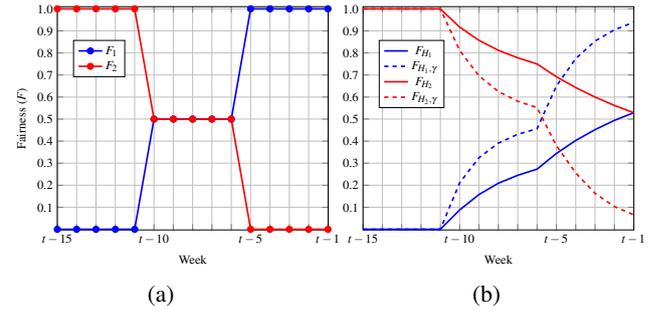
\begin{figure}
    \begin{subfigure}[b]{0.45\columnwidth}
    \centering
    \begin{tikzpicture}[scale=0.52]
    \begin{axis}[
        xmin=-0, xmax=14.01,
        ymin=-.003, ymax=1.01,
        xtick={0,1,2,3,4,5,6,7,8,9,10,11,12,13,14},
        xticklabels={$t-15$,,,,,$t-10$,,,,,$t-5$,,,,$t-1$},
        ytick={0.1,0.2,0.3,0.4,0.5,0.6,0.7, 0.8,0.9, 1.0},
        yticklabels={0.1,0.2,0.3,0.4,0.5,0.6,0.7, 0.8, 0.9, 1.0},,
        grid=both,
        xlabel={Week},
        ylabel={Fairness ($F$)},
        ylabel style={yshift=-10pt},
        legend style={at={(0.25, 0.85)},
                      anchor=north east},
        samples = 100,
        ]
        \addplot[blue, domain=0:14, mark=*,very thick] coordinates {(0,0.000) (1,0.000) (2,0.000) (3,0.000) (4,0.000) (5,0.500) (6,0.500) (7,0.500) (8,0.500) (9,0.500) (10,1.000) (11,1.000) (12,1.000) (13,1.000) (14,1.000)};
        \addplot[red, domain=0:14, mark=*, very thick] coordinates{(0,1.000) (1,1.000) (2,1.000) (3,1.000) (4,1.000) (5,0.500) (6,0.500) (7,0.500) (8,0.500) (9,0.500) (10,0.000) (11,0.000) (12,0.000) (13,0.000) (14,0.000)};
        \legend{$F_1$, $F_2$}
        \end{axis}
    \end{tikzpicture}
    \caption{}
    \label{subfig:sigmoid-history}
    \end{subfigure}
    \quad
    \begin{subfigure}[b]{0.45\columnwidth}
    \centering
        \begin{tikzpicture}[scale=0.52]
    \begin{axis}[
        xmin=-0, xmax=14.01,
        ymin=-.003, ymax=1.01,
        xtick={0,1,2,3,4,5,6,7,8,9,10,11,12,13,14},
        xticklabels={$t-15$,,,,,$t-10$,,,,,$t-5$,,,,$t-1$},
        ytick={0.1,0.2,0.3,0.4,0.5,0.6,0.7, 0.8,0.9, 1.0},
        yticklabels={0.1,0.2,0.3,0.4,0.5,0.6,0.7, 0.8, 0.9, 1.0},,
        grid=both,
        xlabel={Week},
        ylabel style={yshift=-10pt},
        legend style={at={(0.3, 0.85)},
                      anchor=north east},
        samples = 100,
        ]
        \addplot[blue, domain=0:14, very thick] coordinates {(0,0.000) (1,0.000) (2,0.000) (3,0.000) (4,0.000) (5,0.088) (6,0.158) (7,0.209) (8,0.245) (9,0.273) (10,0.344) (11,0.403) (12,0.452) (13,0.494) (14,0.529)};
        \addplot[blue, domain=0:14, dashed, very thick] coordinates {(0,0.000) (1,0.000) (2,0.000) (3,0.000) (4,0.000) (5,0.211) (6,0.325) (7,0.391) (8,0.431) (9,0.456) (10,0.650) (11,0.774) (12,0.854) (13,0.905) (14,0.939)};
        \addplot[red, domain=0:14, very thick] coordinates{(0,1.000) (1,1.000) (2,1.000) (3,1.000) (4,1.000) (5,0.917) (6,0.857) (7,0.812) (8,0.778) (9,0.750) (10,0.692) (11,0.643) (12,0.600) (13,0.562) (14,0.529)};
        \addplot[red, domain=0:14, dashed, very thick] coordinates{(0,1.000) (1,1.000) (2,1.000) (3,1.000) (4,1.000) (5,0.811) (6,0.696) (7,0.625) (8,0.581) (9,0.552) (10,0.379) (11,0.256) (12,0.163) (13,0.103) (14,0.066)};
        \legend{$F_{H_1}$, $F_{H_1, \gamma}$, $F_{H_2}$, $F_{H_2, \gamma}$}
        \end{axis}
    \end{tikzpicture}
    \caption{}\label{subfig:fairness-sigmoid-history}
    \end{subfigure}
\vspace{4mm}
    \caption{Figure \ref{subfig:sigmoid-history} depicts two traces of history with opposite distribution of fairness. The fairness ($F$) of the solution at each point in history is plotted. Figure \ref{subfig:fairness-sigmoid-history} depicts $F_{H}$ and $F_{H, \gamma}$ with $\gamma = 0.65$ for the two given histories in Figure \ref{subfig:sigmoid-history}.}
    \vspace{6mm}
\end{figure}

%
In this section, we evaluate the impact of a history of previous solutions in our proposed approach.
More precisely, we consider two scenarios with equal-length histories, but where the ``distribution'' of the fairness across the histories differs.
By distribution of fairness, we refer to the trend of the fairness of the previous solutions throughout time. 

We consider an instance of NSP with 5 nurses, $n_1, n_2, n_3, n_4$, and $n_5$. At each decision-making time step $t$, we aim to find a solution for a time span of 5 days, where each day has two shifts---morning ($m$) and evening ($e$).
We assume the seniority of the nurses is as follows: $n_1\to 3, n_2\to 2, n_3\to 1, n_4\to 0, n_5\to 0$.
Table~\ref{tab:nurse utility} depicts the utilities assigned by each nurse to the different shifts.
The problem is interesting since the most senior nurses tend to have a stronger preference for morning shifts, whereas our quality metric looks to reward seniority in evening shifts.



We consider two equal-length histories, $H_1$ and $H_2$.
Figure~\ref{subfig:sigmoid-history} depicts the fairness of the solutions of each history.
We observe that the solutions in $H_1$/$H_2$ show an increasing/decreasing trend in fairness across time.
%
%
In fact, $H_1$ and $H_2$ include the exact same solutions, however the order of the solutions is reversed.
Figure \ref{subfig:fairness-sigmoid-history} depicts the value of $F_H$ and $F_{H,\gamma}$ at each time step assuming $\gamma = 0.65$. 
Since $H_1$ and $H_2$ include the same solutions (just in reverse order), it is thus expected that the historical fairness $F_H$ at time step $t - 1$ is the same.
However, when we adopt a discount factor $\gamma = 0.65$, we observe that the value of $F_{H_1,\gamma}$ will be higher than $F_{H_2,\gamma}$.
This is because in $H_1$ ($H_2$) the discount factor starts disregarding the unfair (fair) solutions at the beginning of the history.

\begin{table}
    \centering
        \vspace{2mm}
        \caption{Summary of the solutions generated by \fop , \hfop, and \dhfop with histories $H_1$ and $H_2$. Assumes $\beta = 2$ for all models, and $\gamma=0.65$ for \dhfop.}
        \label{tab:nurse next action}
\renewcommand{\arraystretch}{0.5}
\small
\begin{tabular}{cccccc}
                                       & History & $Q(x_t)$ & $F(x_t)$ \\ \toprule
\multirow{2}{*}{\fop}   & $H_1$   &  $0.09$   &   $0.83$ \\
                                       & $H_2$   &  $0.09$   &   $0.83$                     \\  \midrule
\multirow{2}{*}{\hfop}  & $H_1$   &  $1.0$    &   $0$  
                  \\ 
                                       & $H_2$   &  $1.0$    &   $0$                         \\ \midrule
\multirow{2}{*}{\dhfop} & $H_1$   & $0.63$  &   $0.5$                        \\ 
                                       & $H_2$   &   $0.55$   &  $0$  \\ \bottomrule                       
\end{tabular}
\end{table}
This sets up an interesting experiment, since it allows us to understand the impact of the discount factor in \dhfop.
Table~\ref{tab:nurse next action} summarizes the solutions $x_t$ generated by \fop, \hfop, and \dhfop at time step $t$, under the aforementioned scenario. 
We take $\beta = 2$ for all approaches.
From the results of \fop we observe that this high value of $\beta$ leads to the computation of a lower-quality higher-fairness solution.
We observe as well that \fop generates the same solution in both histories. This is expected, as \fop does not consider previous solutions.
\hfop, however, does take the history of past solutions into account.
Because both histories $H_1$ and $H_2$ included periods of unfair solutions, there still exists a ``fairness debt''.
As a result, \hfop aims to compensate this unfairness by generating an unfair solution. (The solution ends up being high quality as it schedules the senior nurses to go on evening shifts.)
Finally, we have the interesting results of \dhfop.
In $H_1$, due to the discount factor $\gamma$, \dhfop starts disregarding the unfair solutions from earlier time steps.
As a result, there is less ``fairness debt'' to compensate which allowed for a fair solution.
For $H_2$, on the other hand, the discount factor $\gamma$ makes \dhfop disregard the fair solutions from earlier time steps. There is thus a ``fairness debt'' to be compensated, leading to the resulting unfair solution.

In sum, this experiment allowed to assess the impact of the discount factor in \dhfop under histories of previous solutions with different trends of unfairness throughout time.

\begin{table}[t]
\centering
\caption{Utilities assigned by each nurse to the different morning ($m$) and evening ($e$) shifts.}
\vspace{2mm}
\label{tab:nurse utility}
\renewcommand{\arraystretch}{0.3}
\small
\begin{tabular}{ccccccc}
Nurses              & \multicolumn{1}{l}{Shifts} & \multicolumn{1}{l}{Day 1} & \multicolumn{1}{l}{Day 2} & \multicolumn{1}{l}{Day 3} & \multicolumn{1}{l}{Day 4} & \multicolumn{1}{l}{Day 5} \\ \toprule
\multirow{2}{*}{$n_1$} & $m$                          & 3                         & 3                         & 3                         & 3                         & 3                         \\
                    & $e$                          & 0                         & 0                         & 0                         & 0                         & 0                         \\  \midrule
\multirow{2}{*}{$n_2$} & $m$                          & 3                         & 3                         & 3                         & 3                         & 3                         \\
                    & $e$                          & 1                         & 1                         & 1                         & 1                         & 1                         \\  \midrule
\multirow{2}{*}{$n_3$} & $m$                          & 3                         & 3                         & 3                         & 3                         & 3                         \\
                    & $e$                          & 2                         & 2                         & 2                         & 2                         & 2                         \\  \midrule
\multirow{2}{*}{$n_4$} & $m$                          & 0                         & 0                         & 0                         & 0                         & 0                         \\
                    & $e$                          & 3                         & 3                         & 3                         & 3                         & 3                         \\  \midrule
\multirow{2}{*}{$n_5$} & $m$                          & 1                         & 1                         & 1                         & 1                         & 1                         \\
                    & $e$                          & 3                         & 3                         & 3                         & 3                         & 3        \\ \bottomrule               
\end{tabular}
\end{table}

\end{document}